\documentclass[letterpaper, 10 pt, conference]{ieeeconf}  
\usepackage{graphicx}
 \usepackage{multirow}
 \usepackage{amsmath}

\IEEEoverridecommandlockouts                              

\title{\LARGE \bf
Assessment of Input Shaping  for Moving a Wearable Robotic Arm Based on User Comfort and Satisfaction*
}
\title{\LARGE \bf
	User Oriented Assessment of Input Shaping  for Controlling the Vibrations of a Wearable Robotic Arm*
}
\title{\LARGE \bf
	User Oriented Assessment of Command Shaping  as a Vibration Controlling Method in a Wearable Robotic Arm*
}
\title{\LARGE \bf
	User Oriented Assessment of Command Shaping as a Vibration Suppression Method in a Wearable Robotic Arm*
}
\title{\LARGE \bf
	User Oriented Assessment of Vibration Suppression by Command Shaping in a Wearable Robotic Arm*
}
\author{Roozbeh Khodambashi$^{1}$, Gil Weinberg$^{2}$, William Singhose$^{3}$, Shima Rishmawi$^{3}$, Varun Murali$^{3}$, Euisun Kim$^{3}$
\thanks{*This work was supported by National Science Foundation}
\thanks{$^{1}$Roozbeh Khodambashi is with Center for Music Technology,
        Georgia Institute of Technology, 840 McMillan St NW
        Atlanta, GA 30318, USA
        {\tt\small khodambashi@gatech.edu}}%
\thanks{$^{2}$Gil Weinberg is with the Center for Music Technology, Georgia Institute of Technology,
        Atlantaa, GA 30318, USA
        {\tt\small gilw@gatech.edu}}%
\thanks{$^{3}$William Singhose, Sima Rishmawi, Varun Murali and Euisun Kim are with the Department of Mechanical Engineering, Georgia Institute of Technology,
	Atlanta, GA 30313, USA
	{\tt\small singhose@gatech.edu}}%
}
\begin{document}

\maketitle
\thispagestyle{empty}
\pagestyle{empty}

\begin{abstract}
Supernumerary Robotic Limbs (SRLs) exhibit inherently compliant behavior due to the elasticity present at the intersection of human tissue and the robot. This compliance, can prominently influence the operation of some SRLs, depending on the application. In order to control the residual vibrations of SRLs, we have used an input-shaping method which is a computationally inexpensive approach. The effectiveness of this method in controlling the residual vibrations of a SRL has been proven using robustness analysis. User studies show that reducing the vibrations using input shaping directly increases the user satisfaction and comfort by at least 9\%. It is also observed that 36\% of the users preferred unshaped commands. We hypothesize that the shaped commands put a higher cognitive load on the user compared to unshaped commands. This shows that when dealing with human-robot interaction, user satisfaction becomes an equally important parameter as traditional performance criteria and should be taken into account while evaluating the success of any vibration-control method. 
\end{abstract}

\section{INTRODUCTION}
The goal of a supernumerary robotic limb (SRL) \cite{c23},\cite{c24} is to restore or augment human abilities in order to perform tasks that are beyond his/her normal abilities regarding power, speed, and dexterity. SRLs share a common feature, they all have an elastic interface to the human body. Elasticity in the human tissue causes the intersection of the robot and the body to act as an elastic joint. This presents some advantages, such as the possibility to use a multipurpose socket which fits a wider range of body sizes. However, an elastic joint introduces passive compliance in the system. While compliance has been effectively used to increase the safety of robots that operate in close proximity to humans \cite{c20}-\cite{c22}, it results in oscillations. These oscillations affect the quality of physical human-robot interaction. In a human-robot interaction scenario in which the robot arm is attached to the body, these oscillations exert cyclic loads on the user's body which affect the physical comfort of the user. In addition to physical loads, the vibrations put a high cognitive load on the user because robot movements are not predictable, which in turn affects the user's satisfaction. Thus the control of vibrations in compliant robots used as SRLs is critical. 
 
\begin{figure}[t]
	\begin{center}
		\includegraphics[width=8cm,keepaspectratio]{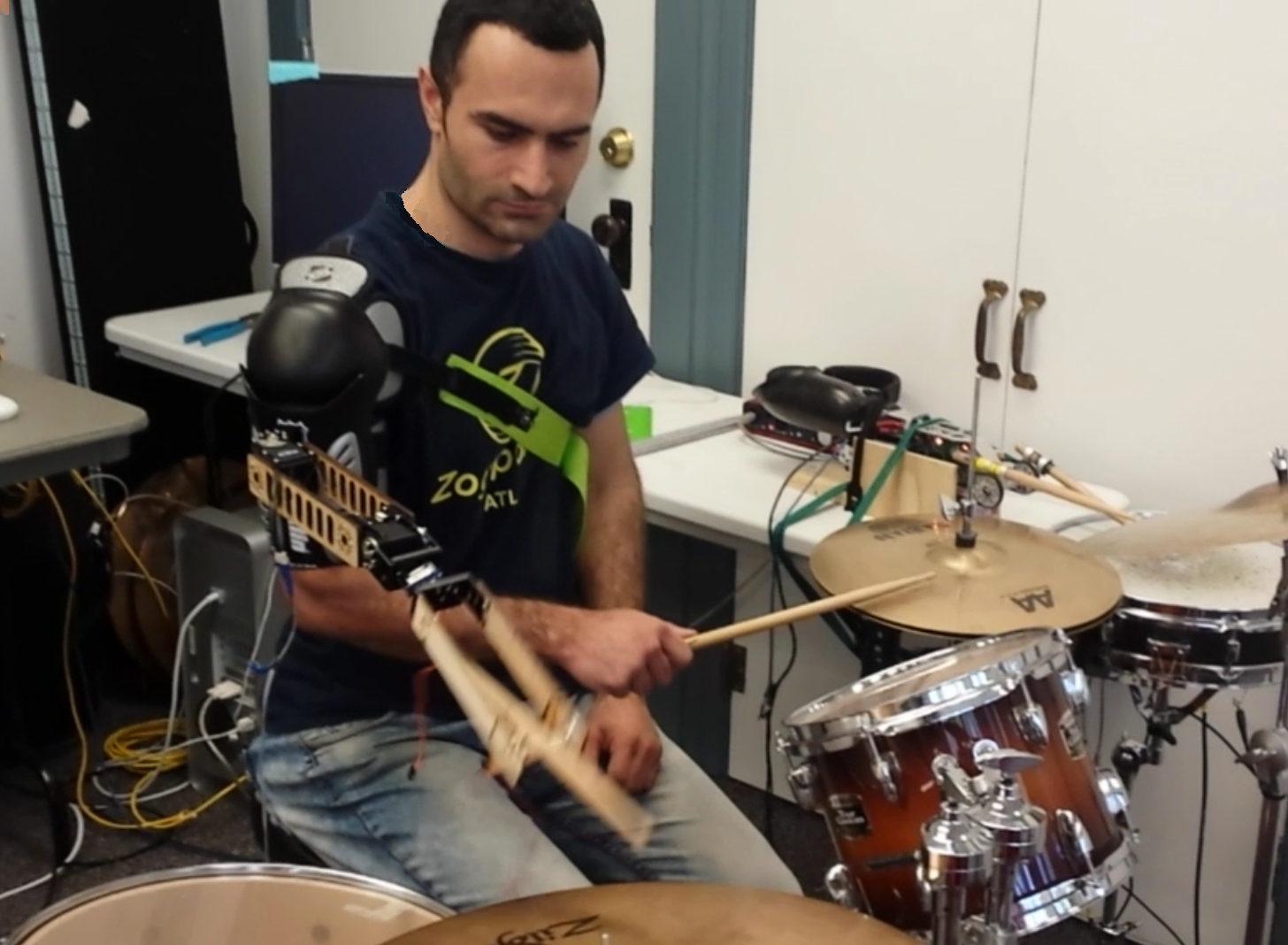}
	\end{center}
	\caption{The 3rdArm platform.}
		\label{fig:thirdarm}
\end{figure}

In this paper, we use input shaping to control the movements of a supernumerary robotic limb. The platform used is the 3rdArm shown in Fig. \ref{fig:thirdarm}. This SRL is attached to the drummer's right shoulder. It moves an additional drum stick to compliment the drummer's own abilities. This platform was designed to study the concept of augmentation and shared control in human-robot interaction. 

Robustness analysis and quantitative measures have been used to evaluate the performance of the control method. In addition, we performed user studies to assess the effectiveness of this control method regarding the user comfort criteria. We found that although input shaping has some advantages, such as simplicity of the algorithm, its performance cannot be evaluated solely based on robustness analysis due to the involvement of a human. 
\subsection{Literature Review}
	Vibration of the elastic members in machines and robotic manipulators is a classical problem which has been well studied. An example of a structure which experiences residual vibrations is the crane in which the position of the end effector (trolley) is not a good estimation of the position of the payload due to inherent swing of the hoist cable. Different methods of controlling residual vibrations of the payload in cranes have been reviewed \cite{c1}. 
	
	Traditionally robots were designed as stiff structures in order to achieve maximum position control even in presence of external disturbances \cite{c25}. This can be dangerous in scenarios which involve physical human-robot interaction (pHRI) because the amount of forces that the robot exerts on the environment cannot be controlled. This problem can be addressed by adding compliance to the robots mechanically or by control techniques. However, this added compliance results in residual vibration of the system. 
	
	A review of the problem of controlling vibrations of robotic manipulators which exhibit compliant behavior due to the presence of flexible joints and links has been presented in \cite{c2}. The goal in almost all these studies is to reduce residual vibrations, improve tracking of inputs, and decrease sensitivity to modelling errors. However, none of these studies have considered this problem from a human-robot interaction perspective. For example, \cite{c3} has studied and compared input shaping and model predictive control (MPC) as two approaches for controlling residual vibrations of a humanoid robot and has concluded that MPC has a superior performance compared to input shaping without considering the impact of using these methods on the user experience.
	
	Comfort is a key factor in the design of wearable robots \cite{c4}-\cite{c8}. The pressure exerted on the skin by the robot is the main parameter that has been known to affect the user comfort \cite{c9}. The authors of \cite{c10} have developed a distributed soft sensor that can measure the pressure distribution in the interaction area without affecting the user comfort. Even the sound generated by the robot may be considered as a parameter that affects the user comfort \cite{c11}. Comfort has different definitions in different wearable technologies and, therefore, various measurement methods have been provided in the literature \cite{c12, c13}. The Quebec User Evaluation of Satisfaction with Assistive Technology (QUEST) \cite{c14} is a widely used tool for measuring the subjective perception of a wearable device \cite{c4}. Another tool is Locally Experienced Discomfort (LED) \cite{c15}.
	
	Recent advances in robotics and the arising need for augmenting the human body has led to a new category of robots called supernumerary robotic limbs (SRLs). The main difference between SRLs and exoskeletons is that SRL are kinematically independent from the human. The limb can move even when the human limb is stationary. SRLs vary in size, shape, material, and the control method. The challenges in the design of SRLs varies based on the application for which they are designed. Smaller SRLs, such as \cite{c16}, might experience fewer vibrations because they have smaller inertia and less actuator power and, therefore, designing a controller that minimizes the vibrations in these SRLs is not necessary. 
	
	The vibrations of the larger compliant robotic arms affect the user comfort in two aspects. First, it exerts loads that are cyclic in nature, which have been shown to alter the tissue in the interface area \cite{c17}. Second, it puts a high cognitive load on the user because the user's concentration is moved from the task at hand, to predicting and correcting the position or trajectory of the robot. 
	
	Human-robot interaction in musical scenarios have been studied previously at the Georgia Tech Center for Music Technology (GTCMT) using different robot platforms such as Shimon \cite{c18}. In order to study how augmenting the human body can affect his performance, the 3rdArm platform was developed which is a SRL attached to the user's shoulder. It helps a drummer perform complicated rhythms, as well as improvise based on his performance. 
	
	The problem of residual vibrations was clearly observable in our experiments with the 3rdArm in which the user paused frequently while playing, looked at the arm and tried to understand its behavior and predict its trajectory and final position in order to make appropriate movements. Therefore, the user often failed to follow the rhythm properly. To address this problem, We have studied these vibrations and used input shaping to control them. In contrary to previous studies, which have not considered user experience in evaluating the performance of vibration-control algorithms, we used a modified and simplified version of QUEST to assess the performance of designed input shaper.
	
	In the next sections, we first describe the 3rdArm platform. Then, we discuss the design and implementation of the input shaper that is used to minimize the residual vibrations of the 3rdArm. Next, we describe the user studies that have been performed to evaluate the performance of the controller regarding comfort criteria. Conclusions are provided at the end.

\section{THE 3RDARM PLATFORM}

\subsection{Physical Description}

The 3rdArm platform is a 4 DOF robotic arm with the capability of attaching to the human body. Fig.~\ref{fig:thirdarm} shows this platform attached to a user's body. The shoulder attachment socket  is made up of a layer of ABS plastic and a layer of soft foam to provide comfort and allow attachment to a wider range of body sizes. Starting from the shoulder mount, the first degree of freedom (DOF) - the shoulder joint - rotates the 3rdArm around the body (horizontal abduction/adduction). The first link connects the shoulder joint to the elbow joint. This joint is the second DOF and performs a function similar to human elbow (flexion/extension). Next is the third DOF which performs rotation of the wrist (supination/pronation). From there, link 2 connects to a fourth DOF which is for moving the drumstick and is solely used for hitting the drum surface. Dynamixel MX-64  servomotors\footnote{$http://www.robotis.com/xe/dynamixel\_en$} are used as actuators. The commands are generated based on musical data in Max/MSP\footnote{$https://cycling74.com/$} software and sent to the motors using an Arduino Mega2560\footnote{$https://www.arduino.cc/en/Main/ArduinoBoardMega2560$} board through serial communication.

\subsection{Dynamics Modeling}
The movements of the second and third DOF do not result in considerable residual vibrations because the effective mass of the system is lower in these cases compared to movements involving the first DOF (shoulder joint). Therefore, for simplicity, we only model the movements of the shoulder joint. We also assume that the second motor is positioned in such a way that link 1 and link 2 are collinear, as in Fig.~\ref{fig:thirdarm}. This is the worst-case scenario in which maximum residual vibration occurs. The arm is moved with maximum actuator effort in order to rotate a distance of $\theta$ in minimum time $t$. This requirement comes from the fact that the robot has to react to the positioning commands as fast as possible or otherwise it cannot play music properly. After the arm is stopped at the end of its travel distance, it continues to vibrate due to the elasticity present in the attachment to the body. 


When dealing with a complicated system, like the robot arm we are trying to control, it is easier to derive the system model using experimental approaches rather than analytical or numerical methods. This is due to many unknown parameters such as the elastic constants and damping ratios of the human tissue, the shoulder mount material, and the robot material. Geometry of the robot also adds to the complexity. To be able to derive a physical model of the arm, its vibrations should be recorded and model parameters should be extracted based on the actual response. To achieve this, a 9 DOF inertial measurement unit (IMU) was mounted at the end of link 2. 
The arm was given an initial displacement and then released to vibrate freely. The angular position of the arm was recorded.

\begin{figure}[t]
	\begin{center}
		\includegraphics[width=8cm,keepaspectratio]{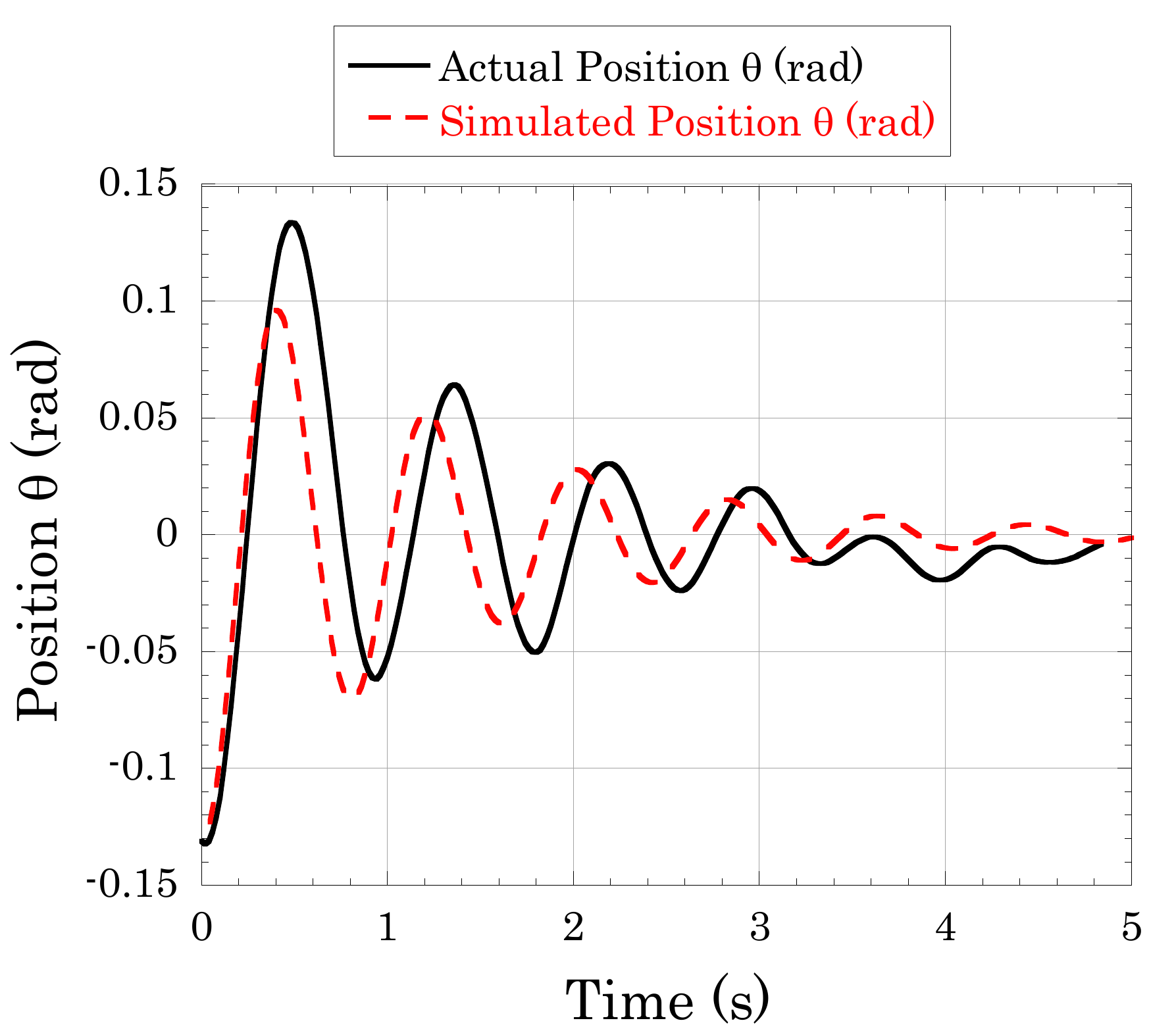}
	\end{center}
	\caption{Actual (recorded) and simulated response of the manipulator to an initial displacement.}
		\label{fig:response}
\end{figure}

By looking at the vibration characteristics of the system due to an initial displacement, which is shown in Fig.~\ref{fig:response} with black solid line, we can see that it is closely matching the response of a simple harmonic oscillator with elastic constant of $K_T$  and damping coefficient of $\zeta$.

The equation of motion of this system is described as:
\begin{equation}
\label{eq:f1}
J\ddot{\theta} +B_T \dot{\theta} +K_T \theta= \tau(t) 
\end{equation}

where: 
\begin{itemize}
	\item $J$ is the rotational inertia (kg$m^2$)
	\item $K_T$ is the torsional spring stiffness (Nm/rad)
	\item $B_T$  is the torsional damping constant (Nms/rad)
	\item $\tau(t)$ is the input torque (Nm)
	\item $\theta$ is the robot arm's rotational displacement (rad)
\end{itemize}

\begin{figure}[t]
	\begin{center}
		\includegraphics[width=8cm,keepaspectratio]{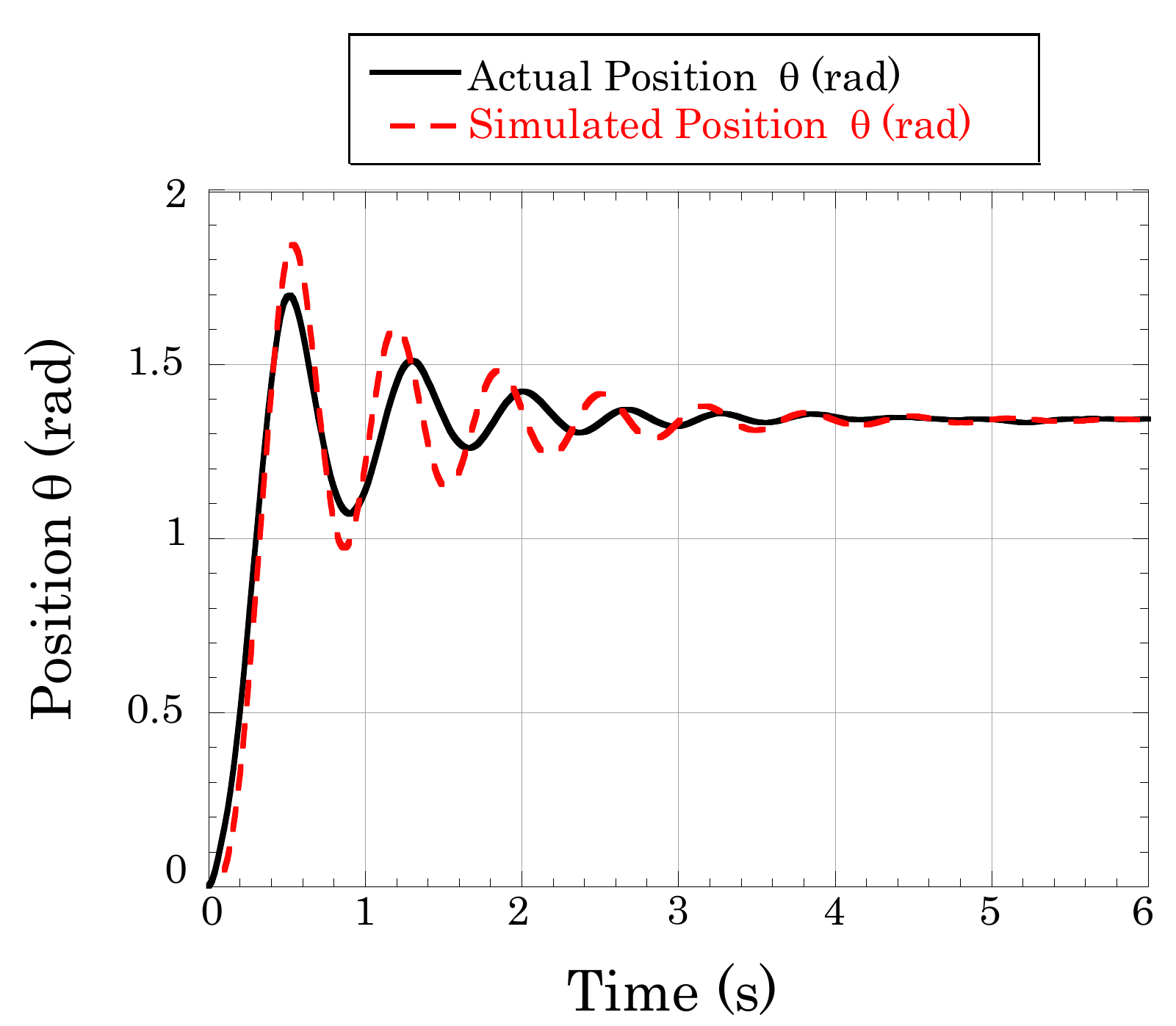}
	\end{center}
	\caption{Actual (recorded) and simulated response of the manipulator to a ramp position input.}
		\label{fig:response-rev}
\end{figure}

Equation~\ref{eq:f1} can be normalized into the following equation:
\begin{equation}
\label{eq:f2}
\ddot{\theta} +2\zeta \omega_n \dot{\theta} +\omega_n^2 \theta=\omega_n^2 u(t)
\end{equation}	

where: 
\begin{itemize}
	\item $\zeta$ is the damping ratio
	\item $\omega_n$  is the natural frequency rad/s
	\item $u(t)$ is the input signal rad
\end{itemize}
Similar behavior can be observed from the response of the system to a ramp position input. This is shown as the black solid line in Fig. \ref{fig:response-rev}.

The damped frequency of oscillations rad/s is calculated using:

\begin{equation}
\omega_d=\frac{2\pi}{T}	
\end{equation}

Where T is the time needed to complete one period of oscillation. Because the damping ratio is relatively small, it can be calculated using logarithmic decrement:
\begin{equation}
\zeta=\frac{\ln \frac{x_0}{x_1}}{2\pi}
\end{equation}

Where $x_0$ and $x_1$ are two successive peaks extracted from the graphs. The natural frequency of oscillation can be calculated by:

\begin{equation}
\omega_n=\frac{\omega_d}{\sqrt{(1-\zeta^2 )}}
\end{equation}

To make sure the calculated parameters are a good approximation of the actual values, both the free and forced responses were recorded five times. The parameters were calculated from all graphs and the average values were found. Results are summarized in Table~\ref{tab:T1}.
\begin{table}[b]
	\centering
	\caption{\textsc{Damping Ratio and Frequency of the System.}}
	\begin{tabular}{|r|r|r|r|}
		\hline
		\multicolumn{1}{|c|}{\multirow{2}[4]{*}{\textbf{Response type}}} & \multicolumn{3}{c|}{\textbf{Calculated parameters}} \\
		\cline{2-4}          & $\omega_d (\mbox{rad/s})$      &  $\zeta$      &  $\omega_n (\mbox{rad}/s)$ \\
		\hline
		Free vibration response & 7.78  & 0.098 & 7.82 \\
		\hline
		Forced vibration response & 9.58  & 0.094 & 9.62 \\
		\hline
	\end{tabular}%
	\label{tab:T1}
\end{table}%
In both cases, the damped and natural frequencies are close because the system experiences a small amount of damping. Also, the oscillation frequencies in the forced vibrations case are greater due to the inner PID feedback controller, which controls the position of the motor, and which comprises only a proportional parameter. The appropriateness of the model is demonstrated by considering the relative similarity of the simulated responses to the experimental responses that are shown in Fig.~\ref{fig:response} and Fig.~\ref{fig:response-rev} with red dashed lines.
\section{Input Shaper Design} 
An input shaper is a sequence of impulses which is convolved with any desired command to create a shaped input that is fed to the system. This will limit residual vibrations. A zero vibration derivative (ZVD) input shaper was selected as the input shaper because it is robust to disturbances and modelling errors, and also is easy to implement. The ZVD shaper was designed based on the calculated system parameters: $\omega_n=9.62$ rad/s and $\zeta=0.1$. 
\subsection{ZVD shaper parameter estimation}
Referring to \cite{c19}, a ZVD shaper consists of three impulses, whose amplitudes and application times are:
\begin{equation}
\label{eq:E6}
\begin{bmatrix}
A_i\\
t_i
\end{bmatrix}
=
\begin{bmatrix}
\frac{1}{(1+k)^2} & \frac{2k}{(1+k)^2} & \frac{k^2}{(1+k)^2}\\
0 & \frac{\pi}{n} & \frac{2\pi}{n}
\end{bmatrix}
\end{equation}
where  $k=e^{\frac{-\zeta\pi}{\sqrt{(1-\zeta^2 )}}}$ . Thus, substituting numerical values in (\ref{eq:E6}), the shaper is expressed by:

\begin{equation}
\label{eq:E7}
\begin{bmatrix}
A_i\\
t_i(s)
\end{bmatrix}
=
\begin{bmatrix}
0.3344 & 0.4877 & 0.1778\\
0 & 0.3266 & 0.6531 
\end{bmatrix}
\end{equation}

\begin{figure}[t]
	\begin{center}
		\includegraphics[width=8cm]{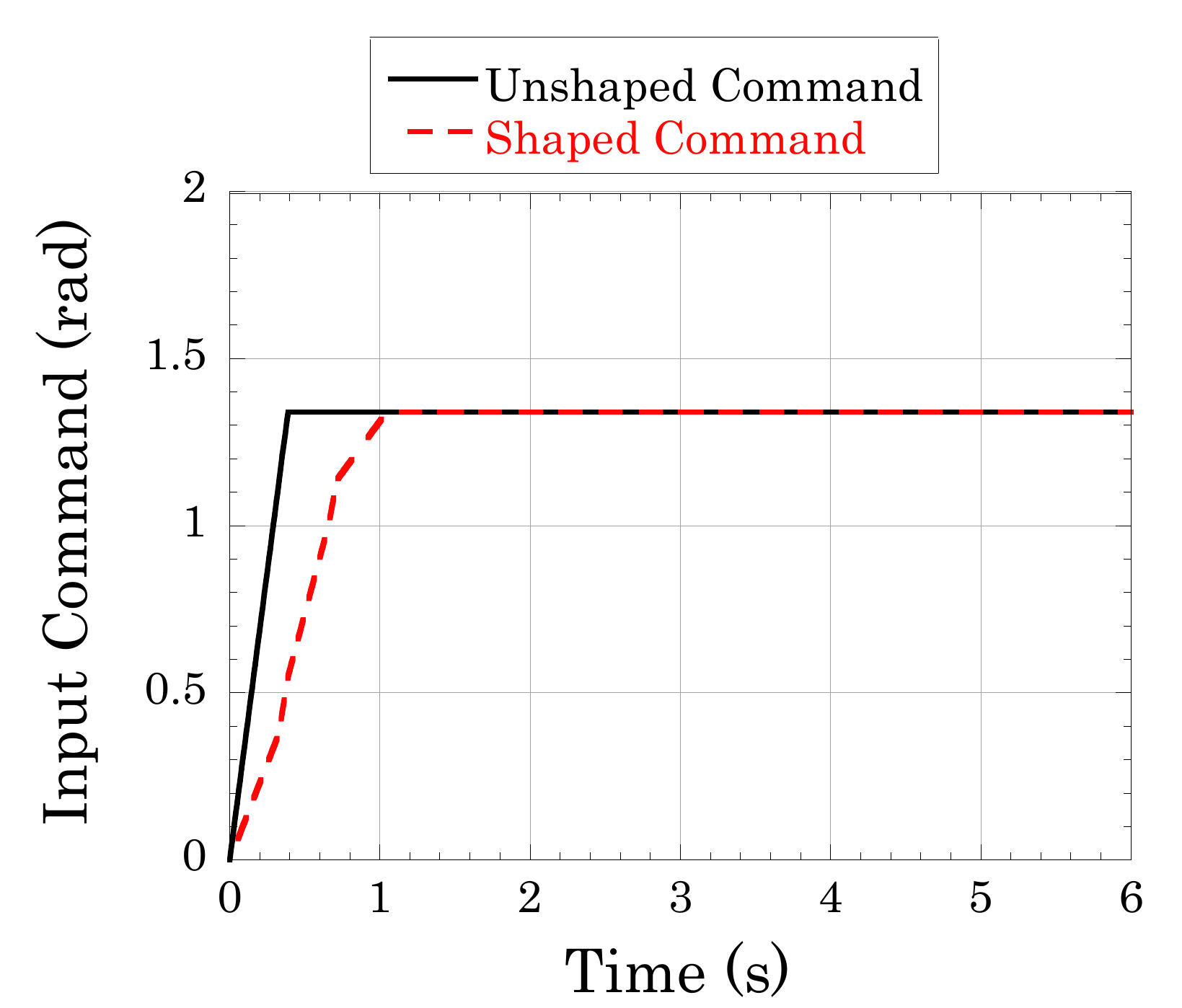}
	\end{center}
	\caption{Unshaped and shaped commands used as input to the system.}
	\label{fig:F5}
\end{figure}

When convolving this shaper with the original input command consisting of a ramp input of $1.45$ rad (solid black line in Fig. \ref{fig:F5}), the result is the shaped input command shown in Fig. \ref{fig:F5} with red dashed line. 

\begin{figure}[t]
	\begin{center}
		\includegraphics[width=8cm]{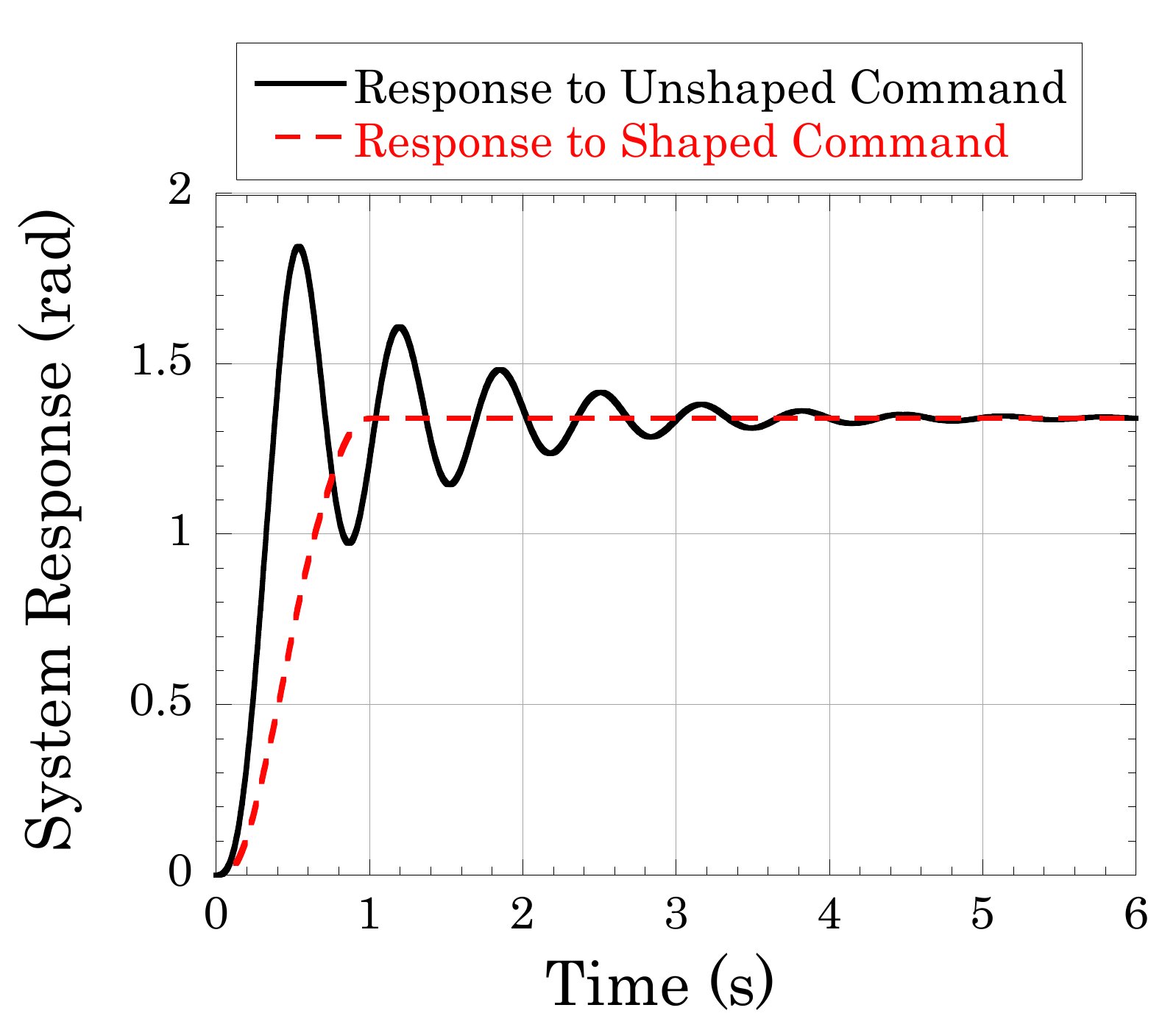}
	\end{center}
	\caption{Simulated unshaped and shaped responses of the system to a ramp input.}
	\label{fig:F7}
\end{figure}

\begin{figure}[t]
	\begin{center}
		\includegraphics[width=8cm]{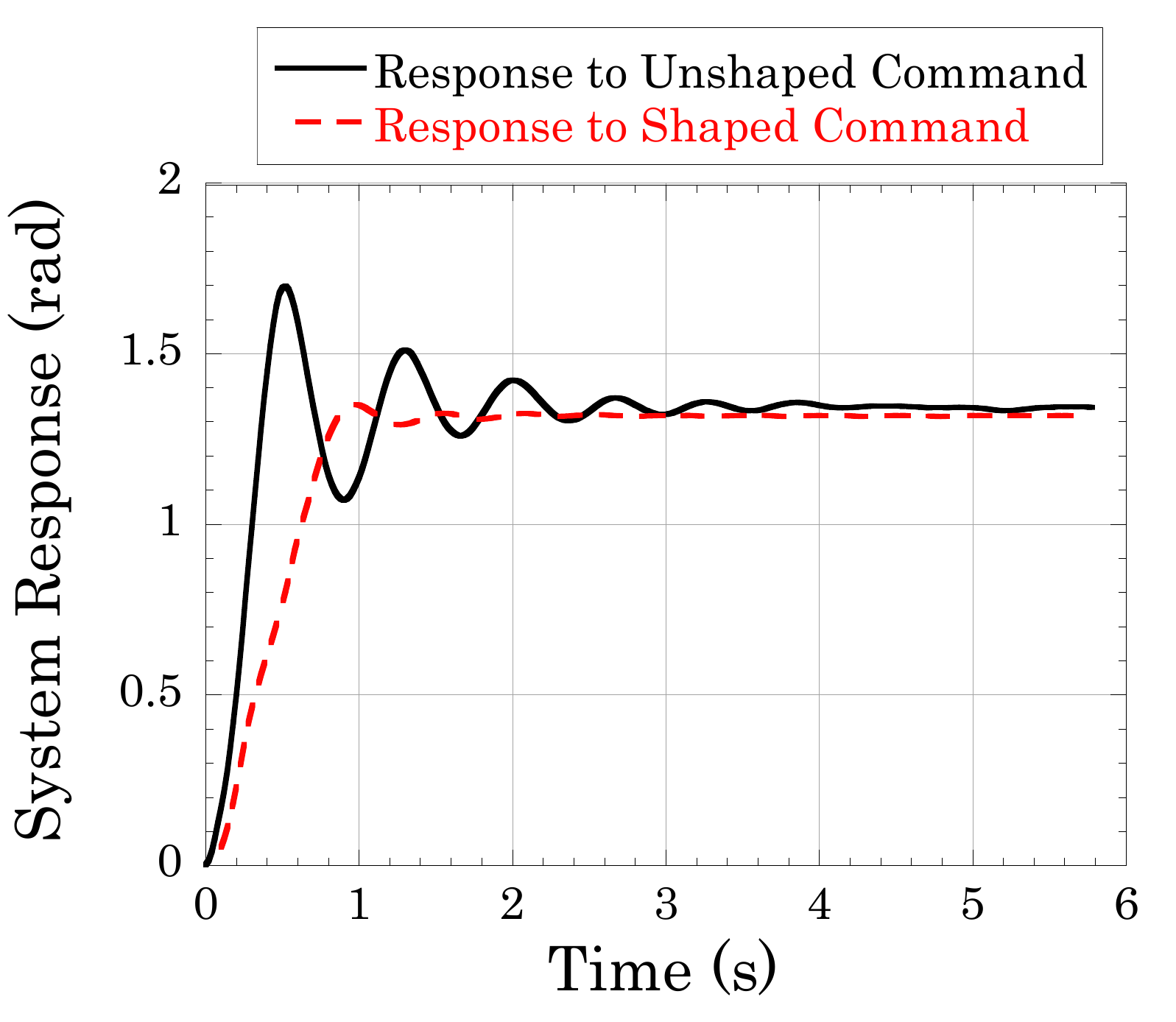}
	\end{center}
	\caption{Actual (recorded) response of the system to a ramp input.}
	\label{fig:F8}
\end{figure}

The simulated system response to the unshaped and shaped commands is shown in Fig.~\ref{fig:F7} with solid black line and red dashed line respectively. It can be noticed that the ZVD input shaper is capable of completely cancelling the residual vibrations. However, the price to be paid here is a time delay of 0.65 s. To obtain the actual response of the system to the shaped commands, a micro-controller was used to divide the original ramp motion profile to five segments, each having a starting and an ending position and a specified speed. This is illustrated in Table~\ref{tab:T2}. In this case, which is illustrated in Fig.~\ref{fig:F8}, the ZVD input shaper was capable of reducing the maximum overshoot to 0.82\%, thus significantly minimizing residual vibrations. Here also, there is a time penalty of 0.6s. This should be accounted for when designing the trajectory of the robot arm while it is playing music.

\begin{table}[b]
	\centering
	\caption{\textsc{Timing and travel angle of each segment in shaped and unshaped commands.}}
	\begin{tabular}{|r|r|r|r|r|}
		\hline
		\multicolumn{1}{|l}{\textbf{Segment}} & \multicolumn{1}{|l}{\textbf{Time Delay (s)}} & \multicolumn{1}{|l}{\textbf{$\theta_i (\mbox{rad})$}} & \multicolumn{1}{|l}{\textbf{$\theta_f(\mbox{rad})$}} & \multicolumn{1}{|l|}{Velocity \textbf{(rad/s)}} \\
		\hline
		\multicolumn{5}{|c|}{\textbf{Unshaped Command}} \\
		\hline
		1     & 0     & 0     & 1.34  & 3.45 \\
		\hline
		\multicolumn{5}{|c|}{\textbf{Shaped Command}} \\
		\hline
		1     & 0     & 0     & 0.37  & 1.154 \\
		2     & 0.32  & 0.37  & 0.57  & 2.84 \\
		3     & 0.4   & 0.57  & 0.96  & 1.68 \\
		4     & 0.62  & 0.96  & 1.15  & 2.3 \\
		5     & 0.73  & 1.15  & 1.34  & 0.62 \\
		\hline
	\end{tabular}%
	\label{tab:T2}%
\end{table}%

\subsection{Robustness Analysis} 
\begin{figure}[t]
	\begin{center}
		\includegraphics[width=8cm]{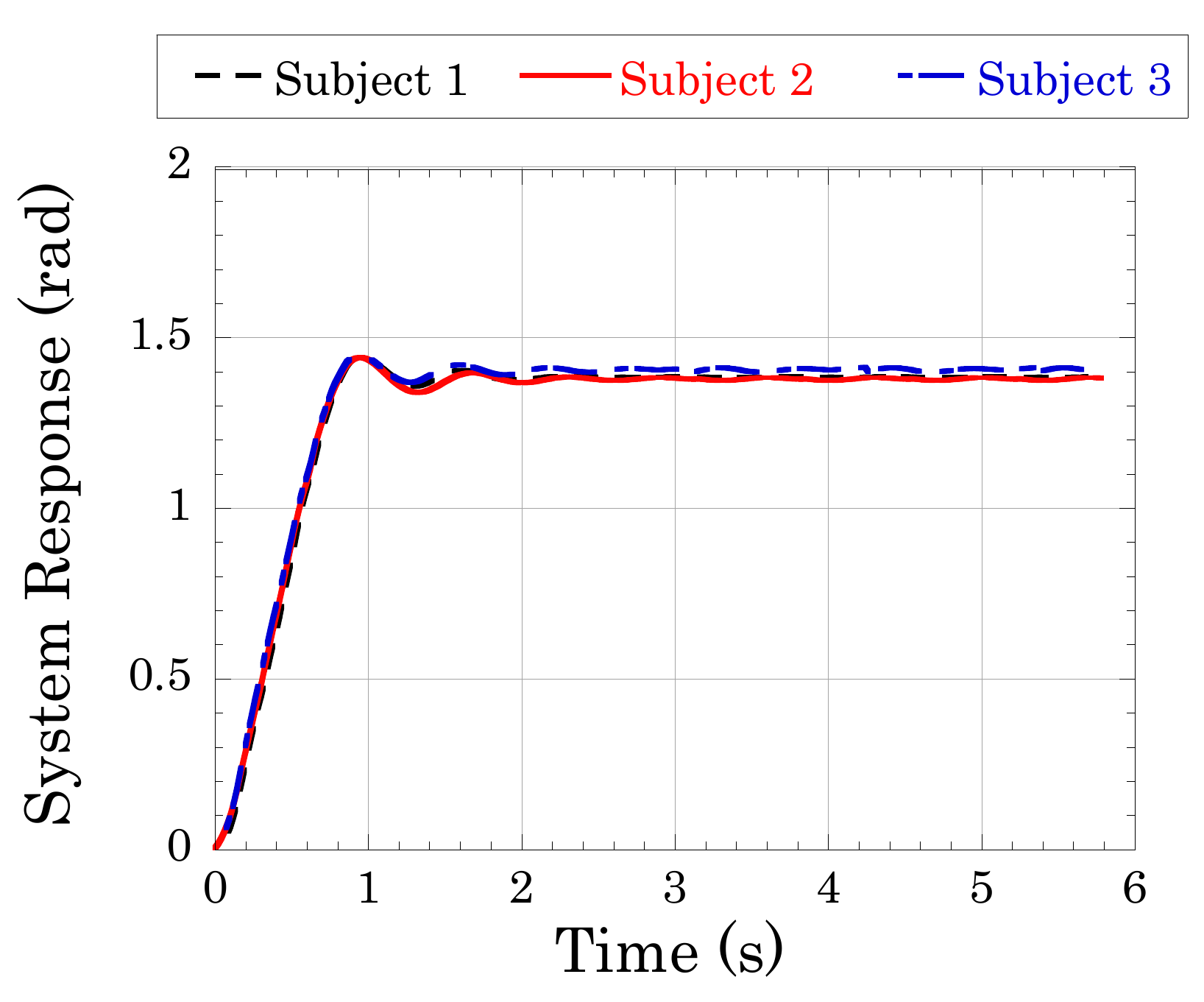}
	\end{center}
	\caption{System ramp response for 3 different users wearing the arm.}
	\label{fig:F9}
\end{figure}

\begin{figure}[t]
	\begin{center}
		\includegraphics[width=8cm]{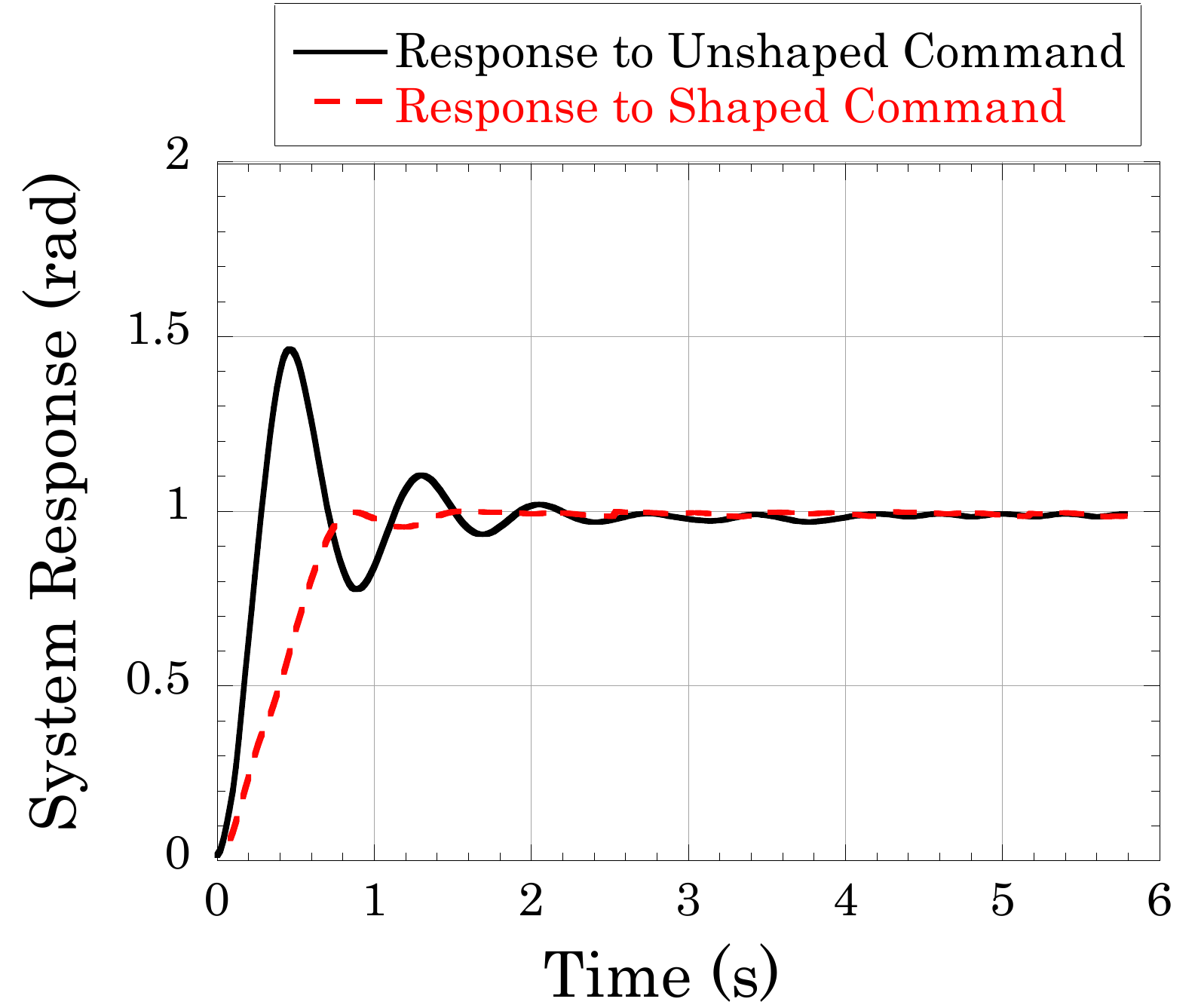}
	\end{center}
	\caption{System response to a move distance of 1 rad.}
	\label{fig:F10}
\end{figure}

Robustness of the input shaper is an important criterion to ensure that the shaper works properly for a wide range of conditions. Two special cases were investigated; first the designed input shaper was tested on different subjects. Every subject had a different arm circumference and a different tissue elasticity. The response of the system for these subjects is shown in Fig.~\ref{fig:F9}. Then, the designed input shaper was used to move the robot arm to a different final desired position (an angular distance of 1 rad) for a single user. The system response is shown in Fig.~\ref{fig:F10}. The data shown in Fig.~\ref{fig:F9} and Fig.~\ref{fig:F10}, demonstrates that the designed input shaper is robust and can be reliably used under different conditions.

\section{USER SURVEY }

   After achieving a successful controller design, we studied the system perfromance on a sample of 14 subjects which were selected through email advertisements and snowball sampling. The study was approved by Georgia Institute of Technology's Institutional Review Board (IRB). Amongst the participants, 57.1\% were male, 64.2\% were between 23-29 years old, 85.7\% played at least one musical instrument and 57.1\% were music technology students. 
   Studies were performed on each participant individually after they completed the consent form and demographics questionnaire.
   
   The circumference and length of the bicep were measured and noted on the questionnaire which were used to observe if there is any correlation between the socket size and the perceived comfort of the 3rdArm movements. Then participants were introduced to the robotic arm. The arm was placed on the participants' shoulder after they confirmed the understanding of the overall process. Three different scenarios were studied. In the first scenario, the arm was kept stationary and the level of comfort with the socket itself was studied. In the second scenario, the 3rdArm was moved using unshaped commands and in the third scenario, shaped command was used. In all the scenarios, the user was sitting stationary on a chair in a comfortable position and observed the movement of the robotic arm. 
   
   A simplified version of the QUEST test was administered in the form of an interview to obtain the perceived comfort of using the robotic arm. This test has 12 satisfaction items which are rated from 1 to 5. Four of these items are related to customer service of the assistive devices which are not considered in our study. From the other 8 items, questions related to weight, dimensions and comfort were relevant to our study and were included in the questionnaire. 
   
\begin{figure}[t]
   	\begin{center}
   		\includegraphics[width=8cm]{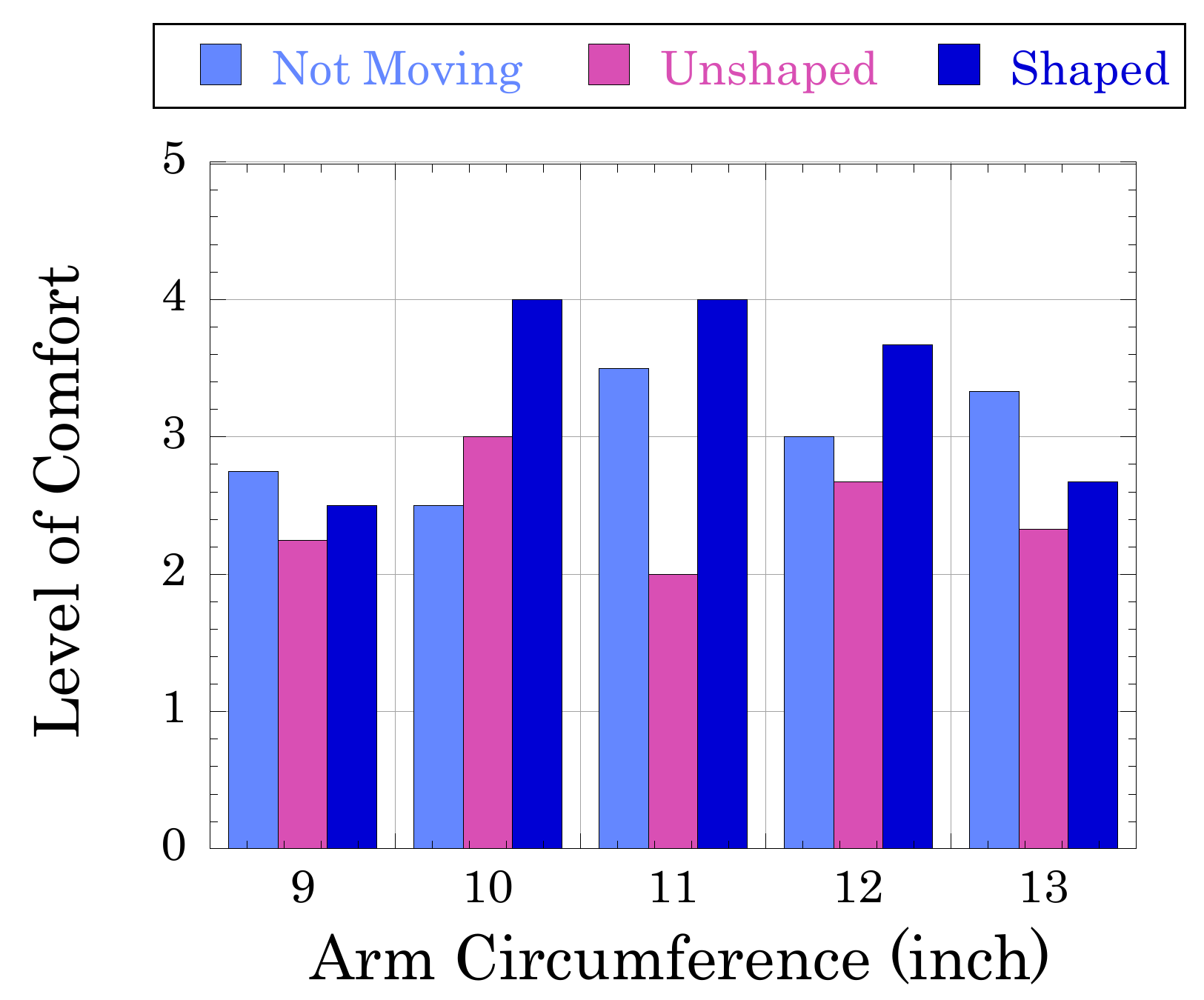}
   	\end{center}
   	\caption{Average user comfort level for arm being stationary, as well as moving with shaped and unshaped commands.}
   	\label{fig:F11}
   \end{figure}

   Fig.~\ref{fig:F11} shows the average level of comfort of the users in three different scenarios based on their arm circumference. The users with arm circumference of 10-11 inches expressed highest comfort. This is due to the fact that the size of the user's arm matched the socket size used for the shoulder attachment. It is also observed that regardless of the arm circumference, the comfort achieved from the shaped command is always higher. 
   
   It was found that although using the input shaper reduces the forces perceived by the user, there is still approximately 36\% of the users that preferred the unshaped commands. This can be attributed to the fact that different users consider different physical and mental criteria for scoring the level of comfort. Users which used physical criteria for judgment mentioned the keywords such as 'less force', 'less recoil' and, 'smoother' in their comments and thus, voted the movement using shaped commands to be more comfortable. On the contrary, users who use the mental parameters in their decision, use keywords such as 'weird' in their comments about the shaped movements and preferred the movements using unshaped commands. This shows that in evaluating the performance of a controller which is used in scenarios involving human-robot interaction, it is important to consider user satisfaction and comfort in addition to commonly used criteria such as input tracking ability and robustness. 
   
\section{CONCLUSIONS}

Experiments with the 3rdArm platform shows that different criteria should be taken into account while designing supernumerary robotic limbs compared to exoskeleton designs or other robotic manipulators. Depending on the application, residual vibrations might be a concern in low impedance robotic limbs including the SRLs. We have shown that the conventional methods for suppressing the residual vibrations in the structures such as cranes and robotic manipulators can be applied to SRLs, effectively the vibrations in them. However, these methods might not increase the user satisfaction and comfort. Therefore, for better judgment about the performance of these methods, user studies have to be taken into account. Future work may include comparing other methods of vibration control such as model predictive control in addition to input shaping to find out the best control strategy based on the level of comfort of the users with these methods.

\addtolength{\textheight}{-12cm}   



%

%



\begin{thebibliography}{99}


\bibitem{c1}	Singhose, W., Command shaping for flexible systems: A review of the first 50 years. International Journal of Precision Engineering and Manufacturing, 2009. 10(4): p. 153-168.
\bibitem{c2}	Dwivedy, S.K. and P. Eberhard, Dynamic analysis of flexible manipulators, a literature review. Mechanism and machine theory, 2006. 41(7): p. 749-777.
\bibitem{c3}	Rupert, L., P. Hyatt, and M.D. Killpack. Comparing Model Predictive Control and input shaping for improved response of low-impedance robots. in Humanoid Robots (Humanoids), 2015 IEEE-RAS 15th International Conference on. 2015.
\bibitem{c4}	Nimawat, D. and P.R.S. Jailiya, Requirement of Wearable Robots in Current Scenario. European Journal of Advances in Engineering and Technology, 2015. 2(2): p. 19-23.
\bibitem{c5}	Mohammed, S., Y. Amirat, and H. Rifai, Lower-limb movement assistance through wearable robots: state of the Art and challenges. Advanced Robotics, 2012. 26(1-2): p. 1-22.
\bibitem{c6}	Gopura, R., et al., Developments in hardware systems of active upper-limb exoskeleton robots: A review. Robotics and Autonomous Systems, 2016. 75: p. 203-220.
\bibitem{c7}	Gálvez-Zúñiga, M.A. and A. Aceves-López, A Review on Compliant Joint Mechanism for Lower-Limb Exoskeletons.
\bibitem{c8}	Gopura, R., K. Kiguchi, and D. Bandara. A brief review on upper extremity robotic exoskeleton systems. in 2011 6th International Conference on Industrial and Information Systems. 2011.
\bibitem{c9}	Rocon, E., et al., Human-Robot Physical Interaction. Wearable Robots: Biomechatronic Exoskeletons, 2008: p. 127-163.
\bibitem{c10}	Lenzi, T., et al., Measuring human-robot interaction on wearable robots: A distributed approach. Mechatronics, 2011. 21(6): p. 1123-1131.
\bibitem{c11}	Veale, A.J. and S.Q. Xie, Towards compliant and wearable robotic orthoses: A review of current and emerging actuator technologies. Medical engineering \& physics, 2016. 38(4): p. 317-325.
\bibitem{c12}	Knight, J.F., et al. The Comfort Assessment of Wearable Computers. in  International Symposium on Wearable Computers (ISWC). 2002.
\bibitem{c13}	Bodine, K. and F. Gemperle. Effects of functionality on perceived comfort of wearables. in Proceedings of the Seventh IEEE International Symposium on Wearable Computers (ISWC'03). 2003.
\bibitem{c14}	Demers, L., R. Weiss-Lambrou, and B. Ska, The Quebec User Evaluation of Satisfaction with Assistive Technology (QUEST 2.0): an overview and recent progress. Technology and Disability, 2002. 14(3): p. 101-105.
\bibitem{c15}	Corlett, E. and R. Bishop, A technique for assessing postural discomfort. Ergonomics, 1976. 19(2): p. 175-182.
\bibitem{c16}	Ort, T., et al. Supernumerary Robotic Fingers as a Therapeutic Device for Hemiparetic Patients. in ASME 2015 Dynamic Systems and Control Conference. 2015. American Society of Mechanical Engineers.
\bibitem{c17}	Mak, A.F., M. Zhang, and D.A. Boone, State-of-the-art research in lower-limb prosthetic biomechanics-socket interface: a review. Journal of rehabilitation research and development, 2001. 38(2): p. 161.
\bibitem{c18}	Hoffman, G. and G. Weinberg. Shimon: an interactive improvisational robotic marimba player. in CHI'10 Extended Abstracts on Human Factors in Computing Systems. 2010.
\bibitem{c19}	Singer, N.C. and W.P. Seering, Preshaping command inputs to reduce system vibration. Journal of Dynamic Systems, Measurement, and Control, 1990. 112(1): p. 76-82.
\bibitem{c20} Bicchi, A., M.A. Peshkin, and J.E. Colgate, Safety for physical human-robot interaction, in Springer handbook of robotics. 2008, Springer. p. 1335-1348.
\bibitem{c21} Bicchi, A., S.L. Rizzini, and G. Tonietti. Compliant design for intrinsic safety: General issues and preliminary design. in Intelligent Robots and Systems, 2001. Proceedings. 2001 IEEE/RSJ International Conference on. 2001.
\bibitem{c22} Park, J.-J., et al. Safe link mechanism based on passive compliance for safe human-robot collision. in Proceedings 2007 IEEE International Conference on Robotics and Automation. 2007.
\bibitem{c23} Llorens-Bonilla, B., F. Parietti, and H.H. Asada. Demonstration-based control of supernumerary robotic limbs. in 2012 IEEE/RSJ International Conference on Intelligent Robots and Systems. 2012.
\bibitem{c24} Parietti, F. and H. Asada, Supernumerary Robotic Limbs for Human Body Support. IEEE Transactions on Robotics, 2016. 32(2): p. 301-311.
\bibitem{c25} Kim, S., C. Laschi, and B. Trimmer, Soft robotics: a bioinspired evolution in robotics. Trends in biotechnology, 2013. 31(5): p. 287-294.

\end{thebibliography}
\end{document}